%% file: main.tex
\documentclass[runningheads]{eccv_2022_llncs}
\usepackage{graphicx}

\usepackage{tikz}
\usepackage{comment}
\usepackage{amsmath,amssymb} 
\usepackage{color}

\usepackage{enumitem}
\usepackage{caption}
\usepackage{multirow}
\usepackage{subcaption}
\usepackage{makecell}
\usepackage{soul}
\usepackage{csquotes}
\usepackage{booktabs}  % Rules in tables
\usepackage{cite}

\usepackage[pagebackref=true,breaklinks=true,letterpaper=true,colorlinks,bookmarks=false,linkcolor={blue}]{hyperref}

\newcommand{\ngaussian}{\mathcal{N}(\mathbf{0}, \mathbf{I})}

\def\crl{CRL}

\newcommand\hmm[1]{\ifnum\ifhmode\spacefactor\else2000\fi>1000 \uppercase{#1}\else#1\fi}
\newcommand{\encname}{\hmm{d}iscrete cosine encoding}
\def\encabbr{DCE}
\def\losabbr{CRL}
\def\methodabbr{\encabbr{}-\losabbr{}}

\newcommand{\R}{\mathbb{R}}

\newcommand{\B}{\mathbf{b}}

\newcommand{\bfh}{\mathbf{h}}

\newcommand{\X}{\mathbf{x}}
\DeclareMathOperator{\relu}{ReLU}

\newcommand{\etal}{\textit{et al}.}
\newcommand{\ie}{\textit{i}.\textit{e}.}
\newcommand{\eg}{\textit{e}.\textit{g}}
\usepackage[accsupp]{axessibility}  % Improves PDF readability for those with disabilities.

\usepackage[width=122mm,left=12mm,paperwidth=146mm,height=193mm,top=12mm,paperheight=217mm]{geometry}

\makeatletter
\@namedef{ver@everyshi.sty}{}
\makeatother
\usepackage{tikz}

\begin{document}
% \renewcommand\thelinenumber{\color[rgb]{0.2,0.5,0.8}\normalfont\sffamily\scriptsize\arabic{linenumber}\color[rgb]{0,0,0}}
% \renewcommand\makeLineNumber {\hss\thelinenumber\ \hspace{6mm} \rlap{\hskip\textwidth\ \hspace{6.5mm}\thelinenumber}}
% \linenumbers
\pagestyle{headings}
\mainmatter
\def\ECCVSubNumber{3}  % Insert your submission number here

\title{Strengthening Skeletal Action Recognizers via Leveraging Temporal Patterns} % Replace with your title

% CAMERA READY SUBMISSION
\titlerunning{DCE-CRL}

\author{Zhenyue Qin$^{1}$\thanks{Email: \textit{zhenyue.qin@anu.edu.au}.}, Pan Ji$^{2}$, Dongwoo Kim$^{3}$, Yang Liu$^{1,4}$, Saeed Anwar$^{1,4}$, \\Tom Gedeon$^{5}$\\
Australian National University$^1$, OPPO US Research$^2$, GSAI POSTECH$^3$, Data61-CSIRO$^4$, Curtin University$^5$\\
}
\authorrunning{Z. Qin et al.}

\maketitle

%%%%%%%%% ABSTRACT
\begin{abstract}
   \input{secs/abs}

\end{abstract}

%%%%%%%%% BODY TEXT
\input{secs/intro}
\input{secs/relater_work/related_work}

\input{secs/methods}

\input{secs/exp/exp}

\input{secs/conclusion}

\clearpage

\bibliographystyle{eccv2022_conference}
\bibliography{egbib}
\end{document}

%% file: secs/abs.tex
Skeleton sequences are compact and lightweight. Numerous skeleton-based action recognizers have been proposed to classify human behaviors. In this work, we aim to incorporate components that are compatible with existing models and further improve their accuracy. To this end, we design two temporal accessories: discrete cosine encoding (\encabbr{}) and chronological loss (CRL). \encabbr{} facilitates models to analyze motion patterns from the frequency domain and meanwhile alleviates the influence of signal noise. CRL guides networks to explicitly capture the sequence's chronological order. These two components consistently endow many recently-proposed action recognizers with accuracy boosts, achieving new state-of-the-art (SOTA) accuracy on two large datasets.

%% file: secs/intro.tex
\section{Introduction}

Accurately recognizing human actions is essential for many applications such as human-robot interaction~\cite{fanello2013keep}, sports analysis~\cite{tran2018closer}, and smart healthcare services~\cite{saggese2019learning}. Recently, skeleton-based action recognition has attracted increasing attention~\cite{2015_cvpr_hrnn,2017_cvpr_c_cnn,2021_tip_shift_gcn}. 
Technically, skeleton data are concise, free from environmental noises (such as background clutter, and clothing), and also lightweight, enabling fast processing speed on edge devices~\cite{2021_tip_shift_gcn}. 
Ethically, skeletonized humans are privacy-protective and racial-unbiased, promoting fair training and inference. 

Recent approaches of skeleton-based action recognition primarily concentrate on devising new network architectures to more comprehensively extract motion patterns from the skeleton trajectories~\cite{2018_aaai_st_gcn, 2020_eccv_decouple_gcn, 2020_cvpr_sgn}.
Most designs focus on proposing various graph operations to capture spatial interactions. On the other hand, the temporal extractors of~\cite{2020_eccv_decouple_gcn,2019_cvpr_dgnn,2020_cvpr_sgn} inherit \cite{2018_aaai_st_gcn}.
In contrast, in this paper, we propose two components that are compatible with existing skeletal action recognizers, to more robustly and adequately extract {\it temporal} movement features. 

\begin{figure}
    \centering
    \includegraphics[width=0.65\linewidth]{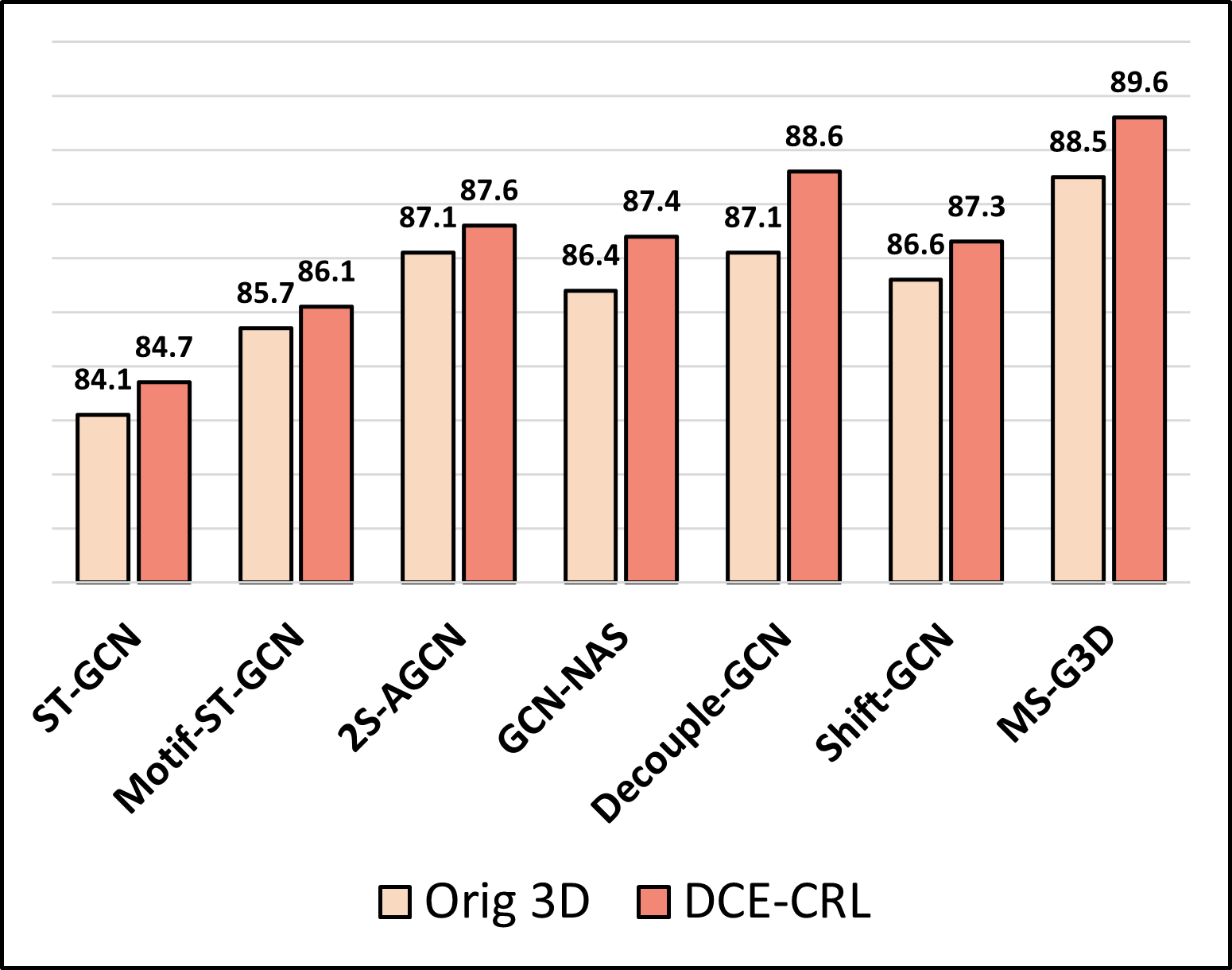}
    \caption{Our proposed \encname{} and chronological loss (\methodabbr{}) consistently improve the accuracy of numerous state-of-the-art models. The results are evaluated on NTU60 with the joint feature. 
    }
    \label{fig:model_accuracy_bar_chart}
\end{figure}

We first present the discrete cosine embedding, which promotes extraction of a motion trajectory's frequency information for a robust skeleton-based action recognizer. To achieve this, we do not explicitly transform the action sequence into the frequency domain; instead, we enrich the sequence by concatenating it with additional encoding. As such, the action recognizer's temporal convolutions on this extra encoding absorb frequency information in the downstream process. 
A primary advantage of the proposed encoding is that it is compatible with most skeleton-based classifiers since most of these models take temporal sequences as input rather than frequency input. Furthermore, this encoding is followed by temporal convolutions of different receptive field sizes, so that frequency features of various temporal resolutions can be effectively learned. Moreover, it also allows for flexible control of the concatenated encoding's frequency. Since temporal noisy pieces of an action (\eg{}, joints in wrongly estimated locations and their sudden absence in some frames) turn out to contain more high-frequency signals than normal parts, we can emphasize the action sequence's low-frequency components to alleviate the noise's adverse influence.

Second, we propose a loss function that facilitates a skeleton-based action recognizer to explicitly capture information about the action sequence's chronological order. The chronological order represents the relative temporal order between two frames, \ie{}, this information depicts whether a frame appears earlier or later than the other. This information offers rich clues, and sometimes is indispensable, for reliably differentiating between actions. For example, it is intuitive that distinguishing between \emph{putting on shoes} and \emph{taking off shoes} is almost impossible without knowing the chronological order of frames.  

We summarize our \textbf{contributions} as follows: 
\begin{enumerate}
\setlength\itemsep{0em}
    \item[i)] We propose discrete cosine embedding. It allows a skeleton-based action recognizer to analyze actions in the frequency domain, and also supports disentangling the signal into different frequency components. We can thus alleviate noise's adverse influence by highlighting low-frequency components. 
    \item[ii)] We present the chronological loss function. It directly guides an action recognizer to explicitly capture the action sequence's temporal information, promoting differentiating temporally confusing action pairs such as \textit{putting on shoes} vs \textit{taking off shoes} and \textit{wearing jacket} vs \textit{taking off jacket}. 
    \item[iii)] The proposed two accessories are widely compatible with existing skeleton-based action recognizers. We experimentally show that the two components consistently boost the accuracy of a number of recent models. 
    \item[iv)] Equipped with these two accessories, a simple model is competitive with more complex models in accuracy and consumes substantially fewer parameters and less inference time. When combining the two components with a more complicated network, we achieve new state-of-the-art accuracy. 
\end{enumerate}

%% file: secs/relater_work/related_work.tex
\section{Related Work}
% \SA{I saw that you have been using the references mostly in the related works. Please mix it up such as sometime put the author name like author~\etal~\cite{2015_cvpr_hrnn} and sometime use the method name such as HRNN~\cite{2015_cvpr_hrnn}. This will make the related work more readible.}
\input{secs/relater_work/skeleton_ar}
\input{secs/relater_work/temp_info}

\input{secs/relater_work/noise_alleviation}

%% file: secs/relater_work/skeleton_ar.tex
\subsection{Recognizing Skeleton-Based Actions}
In early work, the joint coordinates in the entire sequence were encoded into a feature embedding for skeleton-based action recognition~\cite{lei_tip2019}. Since these models did not explore the internal correlations between frames, they lost information about the action trajectories, resulting in low classification accuracy. The advent of convolutional neural networks (CNNs) fortified capturing the correlations between joints and boost the performance of skeleton-based action recognition models~\cite{lei_iccv19,2017_cvpr_c_cnn,liu2017enhanced}. However, CNNs cannot model the skeleton's topology graph, thus missing the topological relationships between joints.

Graph convolution networks (GCNs) were introduced to model these topological relations by representing skeletons as graphs, in which nodes and edges denote joints and bones.
ST-GCN~\cite{2018_aaai_st_gcn} aggregated joint features spatially with graph convolution layers and extracted temporal variations between consecutive frames with convolutions.
In later work, Li \etal{} proposed AS-GCN~\cite{2019_cvpr_as_gcn} to use learnable adjacency matrices to substitute for the fixed skeleton graph, further improving recognition accuracy.
Subsequently, Si \etal{} designed AGC-LSTM~\cite{2019_cvpr_agc_lstm} to integrate graph convolution layers into LSTM as gate operations to learn long-range temporal dependencies in action sequences.
Then, the 2s-AGCN model~\cite{2019_cvpr_2sagcn} utilized bone features and learnable residual masks to increase skeletons' topological flexibility and ensembled the models trained separately with joints and bones to boost the classification accuracy.
Recently, more innovative techniques, such as self-attention~\cite{2020_cvpr_sgn}, shift-convolution~\cite{2020_cvpr_shift_gcn, 2021_tip_shift_gcn} and graph-based dropout~\cite{2020_eccv_decouple_gcn}, have been introduced into GCNs.
% for skeleton-based action recognition.
MS-G3D~\cite{2020_cvpr_msg3d} employed graph 3D convolutions to extract features from the spatial-temporal domain simultaneously. 

%% file: secs/relater_work/temp_info.tex
\subsection{Capturing Chronological Information}
Extracting temporal features has been extensively studied for skeleton-based action recognition. Many of these methods are proposed to more adequately capture motion patterns within a local time window (the adjacent several frames)~\cite{2020_cvpr_msg3d, 2019_cvpr_as_gcn}. Instead, we investigate learning the video-wide temporal evolution. 

Recent skeleton-based action recognition work used the graph neural networks as backbones. They usually did not directly capture the entire motion sequence's relative temporal order. Instead, they gradually increase the sizes of temporal receptive fields to progressively capture temporal information of longer sequences as the networks go deeper. 
In general action recognition, one of the pioneering works that captures the entire video's temporal information was using a linear ranking machine, named rank pooling, to model the temporal order of a video~\cite{2016_cvpr_rank_pooling}. 
Then, Fernando \etal{} proposed the hierarchical rank pooling to more completely capture the action sequence's complex and discriminative dynamics~\cite{2016_cvpr_discriminativer_rp}. At a similar time, \cite{fernando2017discriminatively_ijcv} stated that utilizing subsequences provides more features that are specifically about temporal evolution than merely using single frames. Subsequently, Fernando \etal{} showed how to use deep learning to learn the parameters of rank pooling~\cite{fernando2016learning}. Later, Cherian \etal{} further generalized rank pooling's form with subspaces and formulated the objective as a Riemannian optimization problem~\cite{cherian2018non,cherian2017generalized}. 

The approaches along the direction mentioned above usually modeled learning the video-wide temporal order as an optimization problem. The training was generally not end-to-end with the exception of~\cite{fernando2016learning}. It first transformed the length-varying video to a fixed-length sequence; then, the chronological information was learned on the new representation and required expensively computing the gradients of a bi-level optimization, resulting in slow convergence. 

In this work, based on the spirit of rank pooling, we propose a loss function that facilitates models to explicitly capture the entire sequence-wide chronological information, referred to as the chronological loss.

%% file: secs/relater_work/noise_alleviation.tex
\subsection{Noise Alleviation}
In the literature, work exists that aims to alleviate the noise of skeleton-based action sequences, \eg{}, \cite{2016_eccv_st_lstm} proposed a gate in a recurrent neural network to address noise and \cite{ji2018skeleton} partitioned the skeleton of a human body into different parts and used a part's representation to reduce the noise from a single joint. 

In human motion prediction, Mao \etal{} applied a discrete cosine transform (DCT) to reduce the noise within the motion trajectories~\cite{2019_iccv_mao_dct}. Similarly, our proposed encoding also receives inspirations from~\cite{2019_iccv_mao_dct} and DCTs; however, our approach is different. In~\cite{2019_iccv_mao_dct}, to separate signals of different frequencies, the motion trajectory was firstly mapped to the frequency domain. Then, the graph model processed the signals in the frequency domain. 
In contrast, skeleton-based action recognizers are usually designed to process motion trajectories in the temporal domain.
We evaluated that first converting the signals from the temporal to the frequency domain, followed by using the transformed signals, leads to a plunge of accuracy, reducing more than 10\% according to our experiment. 

We instead propose an encoding mechanism that facilitates disentangling sequences into different frequencies. At the same time, the encoding is still in the temporal domain rather than in the frequency domain. The proposed encodings are thus compatible with many existing models.

%% file: secs/methods.tex
\begin{figure}[t]
\centering
\begin{subfigure}{.26\textwidth}
  \centering
  % include first image
  \includegraphics[width=1.\linewidth]{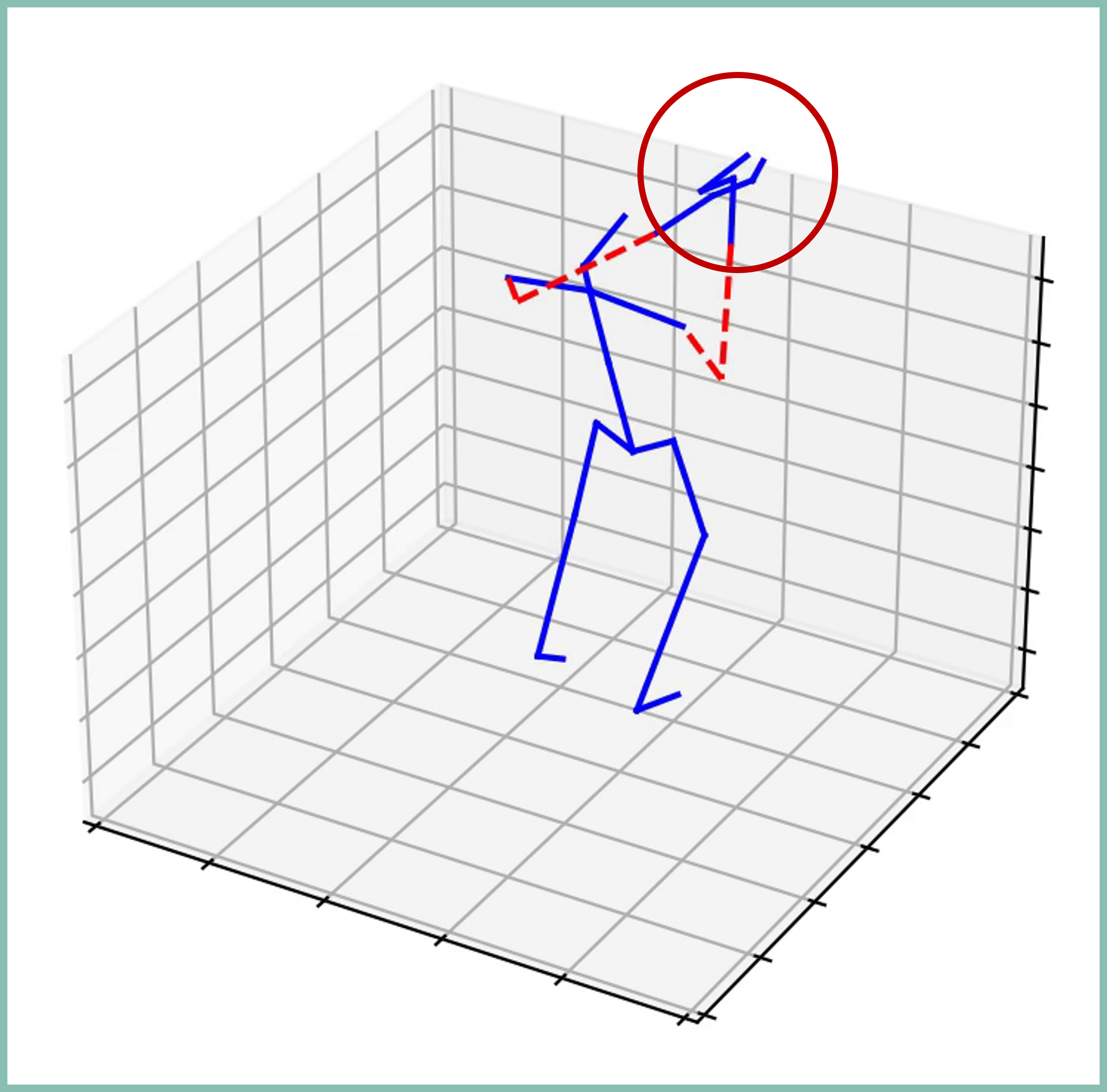}  
  \label{subfig:noise_demo_1}
\end{subfigure}
\begin{subfigure}{.26\textwidth}
  \centering
  % include second image
  \includegraphics[width=1.\linewidth]{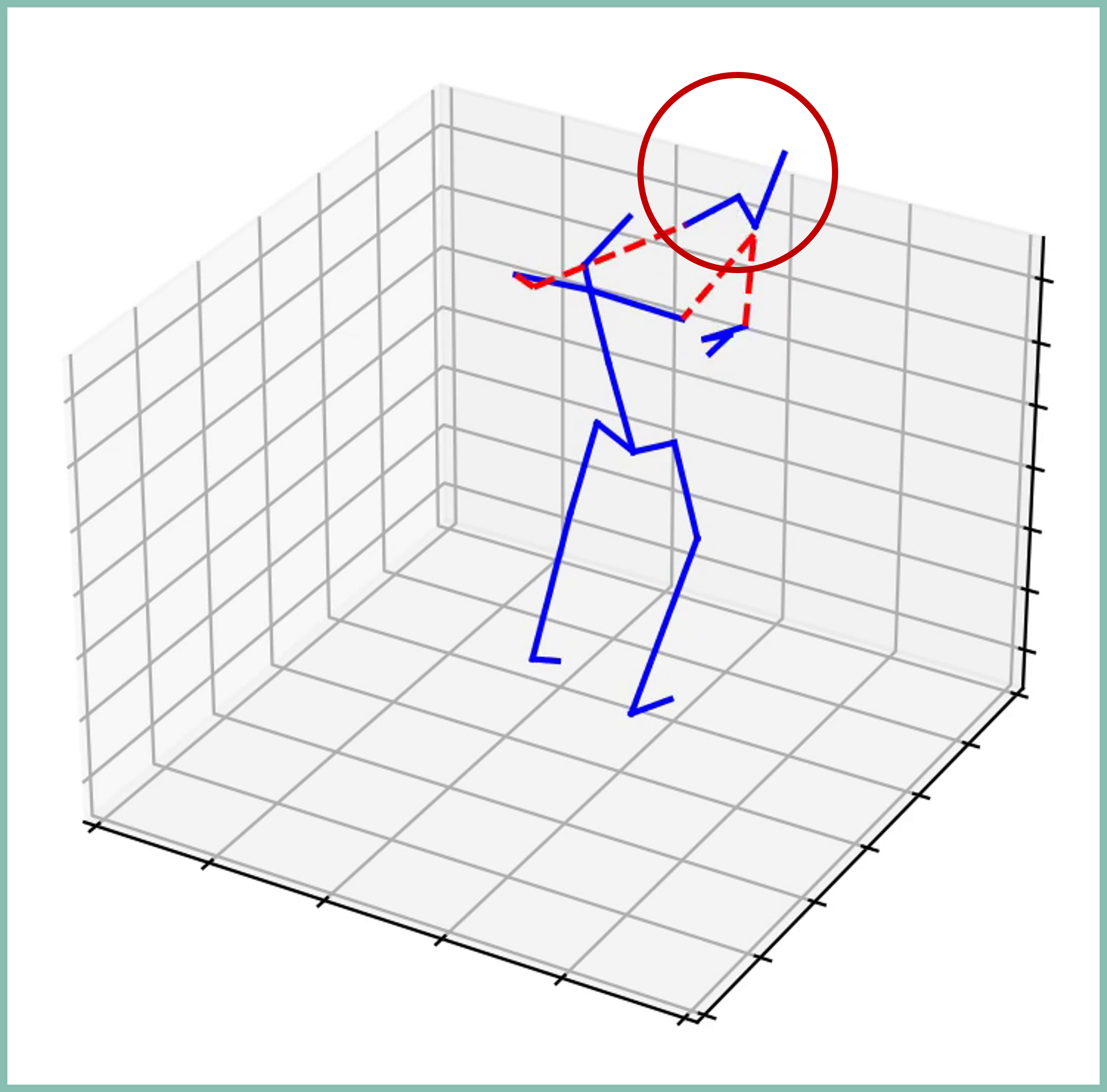}  
  \label{subfig:noise_demo_2}
\end{subfigure}
\begin{subfigure}{.26\textwidth}
  \centering
  % include second image
  \includegraphics[width=1.\linewidth]{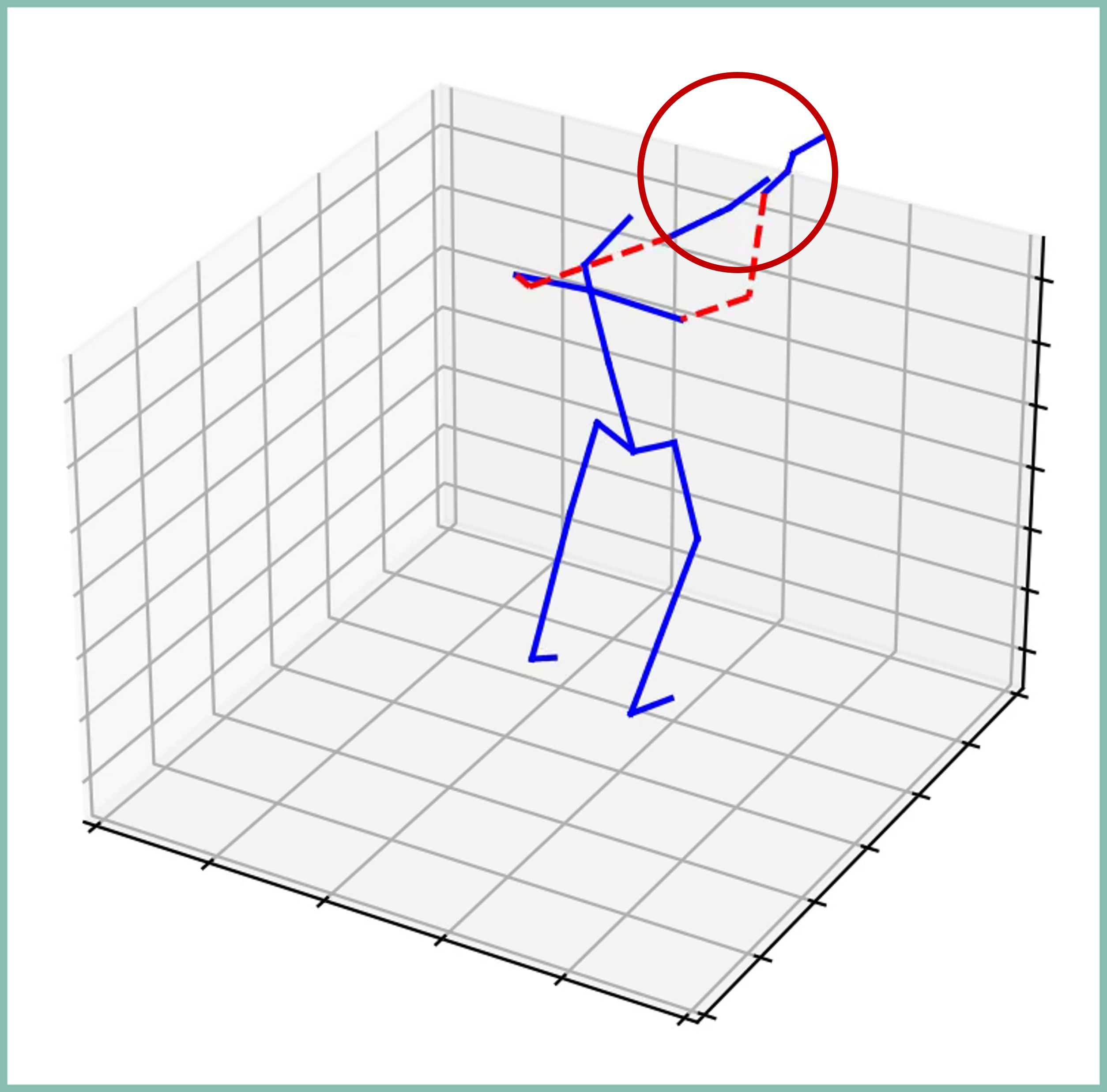}  
  \label{subfig:noise_demo_3}
\end{subfigure}
\vspace{-3mm}
\caption{Illustration of the noisy coordinates highlighted by the red circles. The hand joints show jittering if observing the hand parts of the three figures. In the second figure, the right elbow bends at an unrealistic angle. }
\vspace{-3mm}
\label{fig:noise_demo}
\end{figure}

\begin{figure*}
    \centering
    \includegraphics[width=0.85\linewidth]{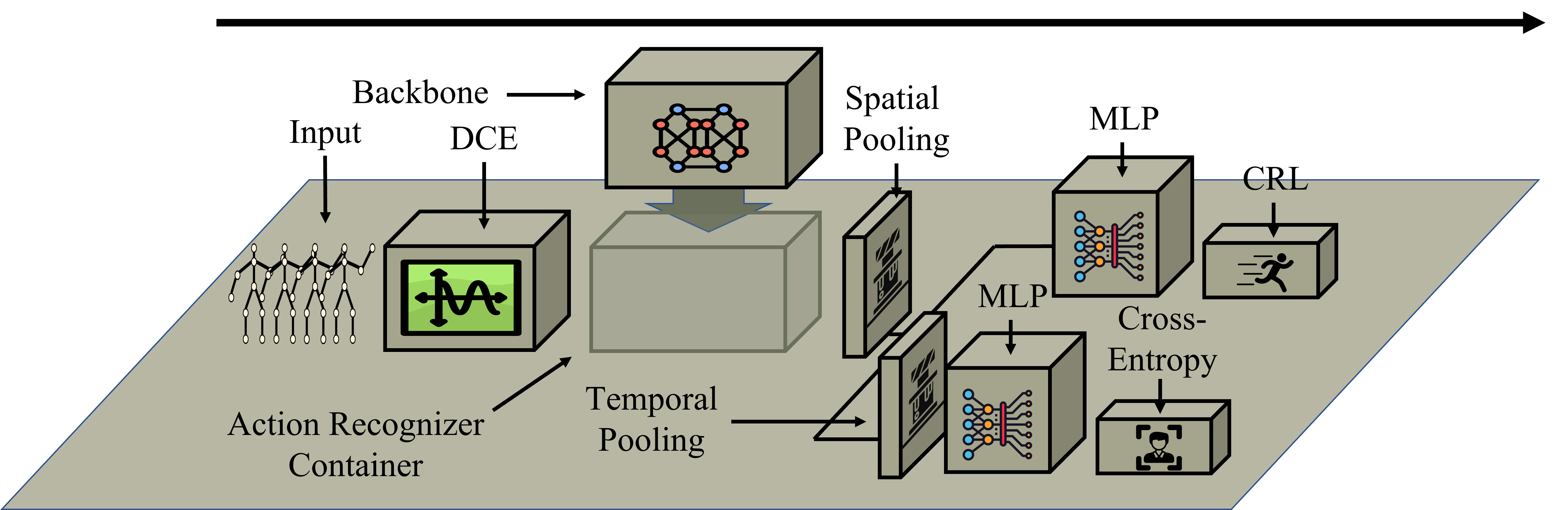}
    \caption{
    Illustration of the framework incorporating the two temporal accessories. Read left to right. 
    The skeleton action sequence is first preprocessed by the \encabbr{} unit. Then it passes through the deep learning backbone, which is flexibly replaceable with most existing skeleton-based action recognizers. The features then undertake spatial pooling. Finally, the features diverge and go through two parallel branches, one for \losabbr{} and the other for classification. 
    }
    \label{fig:framework}
\end{figure*}

\section{Proposed Approach}
Our core motivation is to strengthen the capability of skeleton-based action recognizers to more comprehensively and adequately extract temporal features. To this end, we propose two temporal accessories that can be compatibly inserted into existing models. 
One is the \encname{} that supports mitigating the undesired impact of noise. The other is the chronological loss function that explicitly guides the model in learning relative temporal orders.
In the following, we explain the two components in detail on how they enhance extracting temporal features. 
Then, we describe the framework that simultaneously integrates the two proposed accessories in the end of this section. 

\input{secs/tte/tte}
\input{secs/chron_loss}
\input{secs/framework}

%% file: secs/tte/tte.tex
\input{secs/tte/tte_form}

%% file: secs/tte/tte_form.tex
\subsection{Discrete Cosine Encoding}
In skeleton-based action recognition literature, the input features are usually 3D coordinate-based action movement trajectories along the temporal axis. As in \autoref{fig:noise_demo}, we often observe noise in coordinates, resulting in a wobbling of nodes in the temporal dimension. Action recognizers may unexpectedly misinterpret this as a fast movement, negatively affecting classification accuracy. Although some actions contain fast movements, one can distinguish them from coordinate noise because (i) there is a physical limit to human actions' motion speed, and (ii) noise in coordinates often exhibits oscillating patterns over spatial locations.

If we look closer at the oscillation of spatial locations over consecutive frames, we find that the center of oscillation tends to follow the trajectory of the original action, with a small displacement over the frames. The underlying action trajectory is usually reflected in a low-frequency movement signal. In contrast, the oscillation can generally be viewed as a high-frequency signal that does not affect the actors' main trajectories. We can thus mitigate the noise by suppressing high-frequency signals. 

To this end, we need to decompose the signals into representations over different frequencies. Meanwhile, the decomposed signals remain suitable for existing action recognizers. It is non-trivial to design decomposition approaches that simultaneously satisfy the two requirements. Naively, we may resort to standard methods that map signals in the temporal domain to the frequency domain before feeding them into skeleton-based action recognizers. Nevertheless, three downsides emerge if we follow this path. 

Firstly, the actions like \emph{hopping} and \emph{jumping up} contain fast movements that are intrinsically high-frequency information. Blindly removing all high-frequency components leads to unexpected loss of necessary high-frequency information that are distinct cues for action recognition. To verify this, we mapped original sequences to the frequency domain, and evenly divided the signals into the lower and higher frequencies. Then, we abandoned the higher half and reverted the rest of the signals back to the temporal domain. Compared to the original signals, using these processed sequences dropped the accuracy from 87\% to 72\%.

Secondly, when an action involves more than one person, the temporal co-occurrences between the two peoples' interacting patterns provide rich clues. However, information about temporal locality inevitably reduces as a result of transferring into the frequency domain. Although some approaches like wavelet transform preserve partial temporal locality, the sizes of temporal windows need to be manually determined, which undesirably limits the flexibility of learning comprehensive features.

Thirdly, existing action recognizers are usually \emph{not} designed for processing frequency-domain signals; hence, if we simply feed the frequency signals into such an action recognizer or concatenate them with the original time-domain sequence, the physical meaning will be illogical. 

In sum, we aim to design an encoding mechanism that satisfies two requirements: (i) it supports disentangling high- and low-frequency components of a sequence, enabling mitigating noise; (ii) it can be meaningfully concatenated with the original sequence to prevent the unexpected loss of necessary high-frequency information that are distinct cues for action recognition and ensure compatibility with existing recognizers. 

Our key idea for addressing the above-mentioned three requirements is inspired by the following observation. We notice that many existing action recognizers extract movement features with temporal convolutions. Temporal convolutions conduct weighted sums of features within a local time window. 

Hence, we intend to leverage temporal convolutions to transform the temporal sequences into signals within the frequency domain. The exact procedures are described as follows. We concatenate original sequence $\X = [x_0, \cdots, x_{T-1}]^\top \in \R^{T}$ ($T$ is the frame number) with the following encoding:
\begin{align}
    [\B_0 \odot \X, \cdots , \B_k \odot \X, \cdots, \B_{K-1} \odot \X]\in\mathbb{R}^{K \times T}. 
    \label{eq:dct_cat}
\end{align}
We define $\mathbf{B} = [\B_0, \cdots, \B_k, \cdots, \B_{K-1}] \in \R^{K \times T}$ as a collection of basis sequences. $K$ is a hyperparameter controlling the complexity of $\mathbf{B}$, and $\odot$ represents the Hadamard (element-wise) product. To be more specific, 
the form of a single basis sequence can be written as:
\begin{equation}
    \B_k = \left[ \cos \left( \frac{\pi}{T} (0 + \frac{1}{2}) k \right), \cdots, \cos \left( \frac{\pi}{T} (T-1 + \frac{1}{2}) k  \right) \right]^\top. \notag 
\end{equation}
A larger $k$ corresponds to a basis sequence $\B_k$ with a higher frequency. Thus, selecting $K$ close to $T$ leads to the encoding preserving more high-frequency information of $\X$, and vice versa.

During implementation, $K$ is set to be $8$, 
and the encoding is applied to each joint's 3D coordinates separately and then concatenated with the existing features. Larger $K$ leads to even slightly better performance. A more detailed ablation study on $K$ is in the supplementary materials. 
The shape of the tensor for an action sequence is changed from $\R^{C \times T \times N \times M}$ to $\R^{(K+1)C \times T \times N \times M}$, where $C$ is the feature channel number, $T$ is the sequence length, $N$ is the joint number, $M$ is the subject number. 

One may raise the concern that concatenating with the original sequence seems to preserve the noise. However, our empirical studies show that concatenating with the original signal leads to slightly higher accuracy than discarding the original signal. We conjecture that the network may use the \encabbr{} as references to proactively mitigate the noise of the original sequence. More detailed experimental results are in the supplementary materials. 

The motivation behind applying such an encoding mechanism is to directly guide the action recognizers to easily capture the signal's frequency information. The basis sequence is from the type-2 discrete cosine transform. Hence, a simple summation of the encoding along the temporal direction reveals a sequence's frequency distribution. To facilitate more flexible learning of the frequency information, we replace the summation with a neural network as: 
\begin{align}
    \mathbf{d}_\phi = f_\phi(\B_k \odot \X), 
    \label{eq:d_phi}
\end{align}
where $f_\phi$ is a neural network parameterized by $\phi$. Consequently, $\mathbf{d}_\phi$ can extract frequency-related information on an arbitrary subsequence of $\X$ via temporal convolution. For example, convolution over $\B_k \odot \X$ with a $1 \times 3$ filter can learn how much the signals from three consecutive frames are aligned with a given frequency $k$. If we stack multiple temporal convolution layers, the model can then extrapolate frequency information of longer subsequences. Since temporal convolutional layers are commonly employed in skeleton action recognition models, the encoding can be seamlessly integrated into these models to improve the performance of most existing models. 

We summarize the encoding's advantages as follows:
\begin{enumerate}[leftmargin=*]
\setlength\itemsep{0em}
    \item[i)] The encoding disentangles the action sequence into multiple sequences corresponding to different frequencies. We can mitigate the high-frequency ones to lighten the noise's adverse influence on recognizing actions. 
    \item[ii)] The encoding is in the temporal domain. It thus can be meaningfully concatenated with the original sequence and fed into standard skeletal recognizers. 
    \item[iii)] Since temporal convolutions comprise receptive fields of various sizes, they capture frequency information with various time ranges. 
\end{enumerate}

%% file: secs/chron_loss.tex
\subsection{Chronological Loss Function}
We also aim to design a loss function that enables the action recognizer to correctly capture the chronological order of an action sequence. The order depicts whether a frame appears earlier or later than another frame. To facilitate learning the chronological order, we have to find solutions to the two challenges below: 
\begin{enumerate}[leftmargin=*]
\setlength\itemsep{0em}
    \item[i)] What functions express the relative order between two arbitrary frames?
    \item[ii)] What is the loss function to direct the action recognizer to accurately capture the chronological order? 
\end{enumerate}
In the following, we describe the approaches to tackle the two problems and present the chronological loss that simultaneously resolves the two challenges.  

\textbf{Modeling Relative Order Between Frames}: 
In this part, we explain how to mathematically describe the relative order between two frames. To this end, we require formulating three chronological relationships: a frame appears earlier or later than another frame, or the temporal order does not exist between them. Inspired by~\cite{2016_cvpr_rank_pooling}, we assign each frame at time $t$ with a value $f_\phi(\bfh_t)$, where $f$ is a function (we use a multi-layer perceptron), $\bfh_t$ is the feature of the frame at time $t$, and $\phi$ is the learnable parameter. Then, we use the difference between $f_\phi(\bfh_t)$ and $f_\phi(\bfh_{t'})$ to indicate the temporal order between the frames at time $t$ and $t'$. That is, 
\begin{itemize}[leftmargin=*]
\setlength\itemsep{0em}
    \item $f_\phi(\bfh_i) < f_\phi(\bfh_j)$ iff frame $i$ appears earlier than frame $j$;
    \item $f_\phi(\bfh_i) > f_\phi(\bfh_j)$ iff frame $i$ appears later than frame $j$;
    \item $f_\phi(\bfh_i) = f_\phi(\bfh_j)$ iff the action performer stays idle during the period between frames $i$ and $j$.
\end{itemize}
We refer to the sequence of values:  $$[f_\phi(\bfh_{1}), f_\phi(\bfh_{2}), ..., f_\phi(\bfh_{T})]$$ 
as the chronological values with sequence length $T$. 

\textbf{Formulating the Chronological Loss Function}: 
We have developed a mathematical form that reflects the temporal order between two frames. Now, we aim to devise a loss function that accurately guides the chronological values to indicate the correct temporal order of an action sequence. That is, the chronological values monotonically increase along with the temporal evolution: 
\begin{align}
    \forall t: f_\phi(\bfh_{t}) \le f_\phi(\bfh_{t+1}). 
    \label{eq:chron_obj}
\end{align}

Creating a loss function to guide the model to satisfy the above objective is non-trivial. Poor design causes fluctuations in the chronological values within the sequence. For instance, to let $f_\phi(\bfh_t)$ be smaller than $f_\phi(\bfh_{t+1})$, we may directly add their difference for every two adjacent frames as the loss:
\begin{equation}
    \sum_{t=1}^{T-1} f_\phi(\bfh_t) - f_\phi(\bfh_{t+1}). 
    \label{eq:simply_add_loss}
\end{equation}
However, optimizing the above loss function unintentionally compromises the desired objective specified in~\autoref{eq:chron_obj}. The problem roots in the negative loss when $f_\phi(\bfh_t) \le f_\phi(\bfh_{t+1})$. It can neutralize the positive loss, resulting in undesirably overlooking the cases of $f_\phi(\bfh_t) > f_\phi(\bfh_{t+1})$ for some $t$ within the sequence. 

Motivated by the above failure, we notice that to create a loss function that properly guides the model for fully satisfying the condition outlined in \autoref{eq:chron_obj}, we have to ensure every local decrease of the chronological values is tackled by the loss. To this end, we optimize the loss function in \autoref{eq:simply_add_loss} by leveraging the ReLU function to set the negative loss to zero. The loss function then becomes:
\begin{align}
     \sum_{t=1}^{T-1} \relu \bigg( f_\phi(\bfh_t) - f_\phi(\bfh_{t+1}) \bigg). 
     \label{eq:chron_loss}
\end{align}
As a result, if an earlier frame's chronological value is unintended larger than that of the next frame, a positive loss value will emerge, and not be neutralized. This loss will encourage the model to eliminate the local decrease. The loss value is zero if and only if the chronological values monotonically increase along with the temporal evolutionary direction of the skeleton sequence.

%% file: secs/framework.tex
\subsection{Framework}
The framework incorporating the two temporal accessories is displayed in~\autoref{fig:framework}\footnote{Some icons in the paper's illustrations are from flaticon. }. It is integrally composed of five components. \encabbr{} is the first part to preprocess the skeleton action sequence before feeding it to the second unit for deep learning, which is flexibly replaceable with arbitrary skeleton-based action recognizers. The third piece is a spatial pooling layer. The processing pathway is then divided into two parallel branches to the remaining two components. One is for CRL, and the other is for classification. The dual loss functions are jointly applied during training.

%% file: secs/exp/exp.tex
\begin{table*}[t]\small
\centering
\caption{
Comparison of using \crl{} and \encabbr{} individually and their ensemble. We include the top-1 accuracy and the total number of parameters (\#params) in the networks. \#Ens is the number of models used in an ensemble. 
BSL is baseline. 
Joint and Bone denote the use of joint and bone features, respectively. 
We highlight accuracy improvement with \textbf{\textcolor{magenta}{magenta bold}}. 
} 
\resizebox{0.9\textwidth}{!}{
\begin{tabular}{l c | rrrr | rrrr | r r}
\toprule
&
&
\multicolumn{4}{c| }{\textbf{NTU60}} 
& \multicolumn{4}{c|}{\textbf{NTU120}}
& \multicolumn{1}{c }{\textbf{\# Params}}
& 
\\
% row 2
\cline{3-10}
%\rowcolor{Gray}
\multirow{-2}{*}{\textbf{Methods}} &
\multirow{-2}{*}{\textbf{\# Ens}}  &
X-Sub
&  Acc $\uparrow$
& X-View
&  Acc$\uparrow$ 
& X-Sub
&  Acc$\uparrow$ 
& X-Set
&  Acc$\uparrow$
% & Top-1 & Top-5 
& \multicolumn{1}{c }{\textbf{(M)}}
& \multirow{-2}{*}{\textbf{GFlops}}
\\
\midrule 
BSL (Joint) & 1 & 87.2 & - & 93.7 & - & 81.9 & - & 83.7 & - & 1.44 & 19.4 \\
\crl{} (Joint) & 1 & 88.2 & \textbf{\textcolor{magenta}{1.0}} & 95.2 & \textbf{\textcolor{magenta}{1.5}} & 82.5 & \textbf{\textcolor{magenta}{0.6}} & 84.3 & \textbf{\textcolor{magenta}{0.6}} & 1.77 & 19.6 \\
\encabbr{} (Joint) & 1 & 88.5 & \textbf{\textcolor{magenta}{1.3}} & 95.0 & \textbf{\textcolor{magenta}{1.3}} & 83.1 & \textbf{\textcolor{magenta}{1.2}} & 85.1 & \textbf{\textcolor{magenta}{1.4}} & 1.46 & 19.6 \\
Ens: \methodabbr{} (Joint) & 2 & 89.7 & \textbf{\textcolor{magenta}{2.5}} & 96.0 & \textbf{\textcolor{magenta}{2.3}} & 84.6 & \textbf{\textcolor{magenta}{2.7}} & 86.3 & \textbf{\textcolor{magenta}{2.6}} & 3.21 & 39.2 \\
\midrule
BSL (Bone) & 1 & 88.2 & - & 93.6 & - & 84.0 & - & 85.7 & - & 1.44 & 19.4 \\
\crl{} (Bone) & 1 & 89.4 & \textbf{\textcolor{magenta}{1.2}} & 95.3 & \textbf{\textcolor{magenta}{1.7}} & 84.7 & \textbf{\textcolor{magenta}{0.7}} & 86.6 & \textbf{\textcolor{magenta}{0.9}} & 1.77 & 19.6 \\
\encabbr{} (Bone) & 1 & 90.0 & \textbf{\textcolor{magenta}{1.8}} & 95.1 & \textbf{\textcolor{magenta}{1.5}} & 85.5 & \textbf{\textcolor{magenta}{1.5}} & 86.6 & \textbf{\textcolor{magenta}{0.9}} & 1.46 & 19.6 \\
Ens: \methodabbr{} (Bone) & 2 & 90.7 & \textbf{\textcolor{magenta}{2.5}} & 96.2 & \textbf{\textcolor{magenta}{2.6}} & 87.0 & \textbf{\textcolor{magenta}{3.0}} & 88.3 & \textbf{\textcolor{magenta}{2.6}} & 3.21 & 39.2 \\
\midrule
Ens: BSL (Joint+Bone) & 2 & 89.3 & - & 94.7 & - & 85.9 & - & 87.4 & - & 2.88 & 38.8 \\ 
Ens: \crl{} (Joint+Bone) & 2 & 90.6 & \textbf{\textcolor{magenta}{1.3}} & 96.2 & \textbf{\textcolor{magenta}{1.5}} & 86.1 & \textbf{\textcolor{magenta}{0.2}} & 87.8 & \textbf{\textcolor{magenta}{0.4}} & 3.54 & 39.2 \\
Ens: \encabbr{} (Joint+Bone) & 2 & 90.6 & \textbf{\textcolor{magenta}{1.3}} & 96.6 & \textbf{\textcolor{magenta}{1.9}} & 86.9 & \textbf{\textcolor{magenta}{1.0}} & 88.4 & \textbf{\textcolor{magenta}{1.0}} & 2.92 & 38.9 \\
\bottomrule
\end{tabular}
}
\label{tab:bsl_imp}
\end{table*}

%%%%%%%%%%%%%%%%%%%%%%%
\begin{table}[t]
\caption{
Comparison of action recognition on two benchmark datasets. We compare the recognition accuracy and the total number of parameters (\#params) as well as the inference time (in GFlops) in the networks. 
\#Ens is the number of models used in an ensemble. 
J and B denote the use of joint and bone features, respectively.
BSL represents the utilized backbone model without applying the \methodabbr{}. 
The top accuracy is highlighted in \textbf{\textcolor{red}{red bold}}. 
}
\begin{subtable}{0.49\linewidth}
\centering
% \vspace{-2mm}
\caption{
NTU60. 
} 
\resizebox{1.0\textwidth}{!}{
\begin{tabular}{l c | rrrr | r r}
\toprule
% row 1
%\rowcolor{Gray}
&
&
\multicolumn{4}{c| }{\textbf{NTU60}} 
& \multicolumn{1}{c }{\textbf{\# Para}}
& 
\\
% row 2
\cline{3-6}
%\rowcolor{Gray}
\multirow{-2}{*}{\textbf{Methods}} &
\multirow{-2}{*}{\textbf{\# Ens}}  &
X-Sub
&  Acc $\uparrow$
& X-View
&  Acc$\uparrow$ 
% & Top-1 & Top-5 
& \multicolumn{1}{c }{\textbf{(M)}}
& \multirow{-2}{*}{\textbf{GFlops}}
\\
\midrule 
Lie Group~\cite{2014_cvpr_lie} & 1 & 50.1 & - & 52.8 & - & - & 
\\ 
STA-LSTM~\cite{song2017end} & 1 & 73.4 & - & 81.2 & - & - & - 
\\
VA-LSTM~\cite{2017_iccv_va_lstm} & 1 & 79.2 & - & 87.7 & - & - & - 
\\
HCN~\cite{2018_ijcai_hcn} & 1 & 86.5 & - & 91.1 & - & - & - 
\\ 
MAN~\cite{2018_ijcai_man} & 1 & 82.7 & - & 93.2 & - & - & -
\\ 
ST-GCN~\cite{2018_aaai_st_gcn} & 1 & 81.5 & - & 88.3 & - & 2.91 & 16.4
\\ 
AS-GCN~\cite{2019_cvpr_as_gcn} & 1 & 86.8 & - & 94.2 & - & 7.17 & 35.5
\\
AGC-LSTM~\cite{2019_cvpr_agc_lstm} & 2 & 89.2 & - & 95.0 & - & - & -
\\
2s-AGCN~\cite{2019_cvpr_2sagcn} & 4 & 88.5 & - & 95.1 & - & 6.72 & 37.2
\\
DGNN~\cite{2019_cvpr_dgnn} & 4 & 89.9 & - & 96.1 & - & 8.06 & 71.1
\\
Bayes-GCN~\cite{2019_iccv_bayes} & 1 & 81.8 & - & 92.4 & - & - & -
\\
\midrule 
\multicolumn{8}{c}{Our Method} \\
\midrule 
Ens: \methodabbr{} (J+B) & 2 & \textbf{\textcolor{red}{90.6}} & 8.8 & \textbf{\textcolor{red}{96.6}} & 4.2 & 2.92 & 39.2 \\
\bottomrule
\end{tabular}
}
\end{subtable}
\begin{subtable}{0.49\linewidth}
% \vspace{3mm}
\centering
% \vspace{-2mm}
\caption{
NTU120. 
} 
\resizebox{1.0\textwidth}{!}{
\begin{tabular}{l c | rrrr | r r}
\toprule
% row 1
%\rowcolor{Gray}
&
&
\multicolumn{4}{c| }{\textbf{NTU120}} 
& \multicolumn{1}{c }{\textbf{\# Para}}
& 
\\
% row 2
\cline{3-7}
%\rowcolor{Gray}
\multirow{-2}{*}{\textbf{Methods}} &
\multirow{-2}{*}{\textbf{\# Ens}}  &
X-Sub
&  Acc $\uparrow$
& X-View
&  Acc$\uparrow$ 
% & Top-1 & Top-5 
& \multicolumn{1}{c }{\textbf{(M)}}
& \multirow{-2}{*}{\textbf{GFlops}}
\\
\midrule 
PA LSTM~\cite{2016_cvpr_ntu} & 1 & 25.5 & - & 26.3 & - & - & - \\
Soft RNN~\cite{2018_tpami_soft_rnn} & 1 & 36.3 & - & 44.9 & - & - & - \\
Dynamic~\cite{hu2015jointly} & 1 & 50.8 & - & 54.7 & - & - & - \\
ST LSTM~\cite{2016_eccv_st_lstm} & 1 & 55.7 & - & 57.9 & - & - & - \\ 
% Feature Fusion & 1 & 58.2 & - & 60.9 & - & - & - \\
GCA-LSTM~\cite{liu2017global} & 1 & 58.3 & - & 59.2 & - & - & - \\ 
Multi-Task Net~\cite{2017_cvpr_c_cnn} & 1 & 58.4 & - & 57.9 & - & - & - \\ 
FSNet~\cite{liu2019skeleton} & 1 & 59.9 & - & 62.4 & - & - & - \\ 
Multi CNN~\cite{2018_tip_representation} & 1 & 62.2 & - & 61.8 & - & - & - \\ 
Pose Evo Map~\cite{liu2018recognizing} & 1 & 64.6 & - & 66.9 & - & - & - \\ 
SkeleMotion~\cite{caetano2019skelemotion} & 1 & 67.7 & - & 66.9 & - & - & - \\
2s-AGCN~\cite{2019_cvpr_2sagcn} & 1 & 82.9 & - & 84.9 & - & 6.72 & 37.2 \\
\midrule 
\multicolumn{8}{c}{Our Method} \\
\midrule 
Ens: \methodabbr{} (J+B) & 2 & \textbf{\textcolor{red}{87.0}} & 4.1 & \textbf{\textcolor{red}{88.4}} & 3.5 & 2.92 & 39.2 \\
\bottomrule
\end{tabular}
}
\end{subtable}
\label{tab:compare_with_sota}
\end{table}

\begin{table*}[t!]
\caption{
Comparison with and without the \methodabbr{} on the confusing actions. 
The \enquote{Action} column shows the ground truth labels, and the \enquote{Similar Action} column shows the predictions from the model (with/without the \encabbr{} and/or CRL).
The accuracy improvements highlighted in \textbf{\textcolor{red}{red}} are the substantially increased ones (Acc$\uparrow$ $\geq$ 5\%) due to using the \encabbr{} and/or CRL. 
The similar actions in \textbf{\textcolor{cyan}{cyan}} show the change of predictions after employing \encabbr{} and/or CRL.
}
\label{table:class_improvement}
\begin{subtable}{\linewidth}
\centering
\caption{Examples of actions whose confusing ones temporally evolve oppositely. }
\resizebox{0.9\textwidth}{!}{
\begin{tabular}{l|rc|rrc}
\toprule 
\multirow{2}{*}{\textbf{Action}} & \multicolumn{2}{c|}{\textbf{Joint}} & \multicolumn{3}{c}{\textbf{Joint with \encabbr{}}} \\
\cline{2-6}
& Acc (\%) & Similar Action & Acc (\%) & Acc$\uparrow$ (\%) & Similar Action \\
\midrule 
putting on bag & 91.0 & \textbf{\textcolor{cyan}{taking off jacket}} & 93.9 & 2.9 & \textbf{\textcolor{cyan}{wearing jacket}} \\
taking off a shoe & 81.0 & wearing a shoe & 83.2 & 2.2 & wearing a shoe \\ 
putting on headphone & 85.6 & \textbf{\textcolor{cyan}{taking off headphone}} & 87.7 & 2.1 & \textbf{\textcolor{cyan}{sniff}} \\ 
\bottomrule 
\end{tabular}
}
\label{subtable:CRL_confuse}
\end{subtable}
\begin{subtable}{\linewidth}
\centering
\caption{Actions whose accuracy gets most improved by the CRL. }
\resizebox{0.9\textwidth}{!}{
\begin{tabular}{l|rc|rrc}
\toprule 
\multirow{2}{*}{\textbf{Action}} & \multicolumn{2}{c|}{\textbf{Joint}} & \multicolumn{3}{c}{\textbf{Joint with CRL}} \\
\cline{2-6}
& Acc (\%) & Similar Action & Acc (\%) & Acc$\uparrow$ (\%) & Similar Action \\
\midrule 
opening a box & 66.4 & folding paper & 73.9 & \textbf{\textcolor{red}{7.5}} & folding paper \\ 
making a phone call & 78.2 & playing with phone & 84.4 & \textbf{\textcolor{red}{6.2}} & playing with phone \\
playing with phone & 60.0 & stapling book & 64.7 & 4.7 & stapling book \\
typing on a keyboard & 64.4 & writing & 69.1 & 4.7 & writing \\
balling up paper & 65.7 & folding paper & 69.9 & 4.2 & folding paper \\
\bottomrule 
\end{tabular}
}
\label{subtable:CRL_improve}
\end{subtable}
\begin{subtable}{\linewidth}
\centering
\caption{Actions whose accuracy gets most improved by the \encabbr{}. }
\resizebox{0.9\textwidth}{!}{
\begin{tabular}{l|rc|rrc}
\toprule 
\multirow{2}{*}{\textbf{Action}} & \multicolumn{2}{c|}{\textbf{Joint}} & \multicolumn{3}{c}{\textbf{Joint with \encabbr{}}} \\
\cline{2-6}
& Acc (\%) & Similar Action & Acc (\%) & Acc$\uparrow$ (\%) & Similar Action \\
\midrule 
yawn            & 63.5 & hush (quiet)          & 72.2 & \textbf{\textcolor{red}{8.7}} & hush (quiet)          \\
balling up paper   & 65.7 & \textbf{\textcolor{cyan}{folding paper}}            & 73.9 & \textbf{\textcolor{red}{8.2}} & \textbf{\textcolor{cyan}{playing magic cube}}        \\
wipe            & 83.0 & touching head & 90.9 & \textbf{\textcolor{red}{7.9}} & touching head \\
playing magic cube & 61.7 & \textbf{\textcolor{cyan}{playing with phone}}        & 68.7 & \textbf{\textcolor{red}{7.0}} & \textbf{\textcolor{cyan}{counting money}}         \\
making ok sign    & 38.1 & making victory sign     & 44.0 & \textbf{\textcolor{red}{5.9}} & making victory sign    \\
\bottomrule 
\end{tabular}
}
\label{subtable:TTE_improve}
\end{subtable}
\end{table*}

\section{Experiments}
\input{secs/exp/datasets}
\input{secs/exp/setups}
\input{secs/exp/backbone}
\input{secs/exp/ablation}

\input{secs/exp/baseline}
\input{secs/exp/action_imp}

\input{secs/exp/noise_alleviation}

\input{secs/exp/compatibility}
\input{secs/exp/sota}

%% file: secs/exp/datasets.tex
\subsection{Datasets} 

\noindent
\textbf{NTU60}~\cite{2016_cvpr_ntu}.
As a widely-used benchmark skeleton dataset, NTU60 contains 56,000 videos collected in a laboratory environment by Microsoft Kinect V2. This dataset is challenging because of different viewing angles of skeletons, various subject skeleton sizes, varying speeds of action performance, similar trajectories between several actions, and limited hand joints to capture detailed hand actions. NTU60 has two subtasks, \ie{}, cross-subject, and cross-view. For cross-subject, data of half of the subjects are used for training, and those of the rest are for validation. For the cross-view task, various cameras are placed in different locations, half of the setups are to train, and the others are to validate. 

\noindent
\textbf{NTU120}~\cite{2019_tpami_ntu120}: 
An extension of NTU60 with more subjects, more camera positions and angles, resulting in a more challenging dataset with 113,945 videos.

%% file: secs/exp/setups.tex
\subsection{Experimental Setups}
Our models employ PyTorch as the framework and are trained on one NVIDIA 3090 GPU for 60 epochs. Stochastic gradient descent (SGD) with momentum 0.9 is used as the optimizer. The learning rate is initialized as 0.05 and decays to be 10\% at epochs 28, 36, 44, and 52. In the footsteps of \cite{2019_cvpr_dgnn}, we preprocess the skeleton data with normalization and translation, padding each video clip to be of the same length of 300 frames by repeating the actions. Models are trained independently for NTU60 and NTU120. 

Our \methodabbr{} is generally compatible with existing models. We choose a modified MS-G3D~\cite{2020_cvpr_msg3d} as our baseline architecture for studying the proposed methods. We omit the G3D module because 3D graph convolutions are time-consuming, doubling the training time if employing it, according to our empirical evaluation. We refer to this simplified model as BKB, standing for the backbone. More details about BKB are in the supplementary materials. We will release the code once the paper is published. 

%% file: secs/exp/backbone.tex
% \subsection{Backbone}
% We apply a simplified MS-G3D~\cite{2020_cvpr_msg3d} as the backbone of our model. The simplicity lies on the removal of the computationally heavy G3D component. More details are in the supplementary materials. 

% Empirically, this leads to the inference time to become 40\% of the original. The backbone's illustration is pictorially depicted in \autoref{fig:backbone}. The skeleton sequence sequentially passes through three spatial-temporal units (STUs), as exhibited in \autoref{fig:backbone}(a). An STU consists of a spatial feature extractor (SFE) and three temporal pattern extractors (TPEs). 

%% file: secs/exp/ablation.tex
\subsection{Ablation Studies}
In \autoref{tab:bsl_imp}, we report the results of evaluating the influence of \losabbr{} and \encabbr{}. The assessments include applying the two components individually to compare against training without the proposed methods. Existing approaches reveal great accuracy improvement resulting from ensembling multiple models~\cite{2020_cvpr_msg3d,2020_cvpr_shift_gcn,2020_eccv_decouple_gcn}. Hence, we also evaluate ensembling two models trained with the components separately. 

We have three observations: 
First, both \losabbr{} and \encabbr{} substantially improve the baseline action recognizer's accuracy over all four datasets. For most cases, \encabbr{} enhances an action recognizer more than CRL, reflected in the higher improved accuracy. This implies that \encabbr{} may provide more complementary information to the recognizer. 
% Bones improves more than joints. 
Second, previous work indicates that using the bone features leads to higher accuracy than applying the joint coordinates. We discover that both CTR and \encabbr{} improve accuracy more in the bone domain than the joint domain, implying that bones are not only more suitable representations for skeleton-based action recognition, but also they have richer potential features to be further exploited.
Third, ensembling the two models trained with CRL and \encabbr{} further largely boosts the accuracy, suggesting that CRL and \encabbr{} provide complementary information to each other. We conjecture that CRL focuses on capturing the chronological order, while \encabbr{} alleviates the noise. 

% Dimensionality 
One may assume that \encabbr{}'s boost is due to the dimension increase rather than the encoding. To answer this, we also evaluate the BKB's accuracy using other two approaches. These two extend the dimensionality to be the same as the sequence with the \encabbr{}: 1) The extended dimensional values are the elementwise product of the randomly generated values between $-1$ and $1$ and the original sequence; the range between $-1$ and $1$ coincides with the cosine function's amplitude. 2) The enlarged dimensions are the simple repetition of the original skeleton 3D coordinates. 

We observe that feeding either of the above two inputs to the BKB consequently leads to an accuracy decrease. Using the joint features of the NTU120 cross-subject setting, the accuracy slightly drops from 81.9\% to 80.7\% and 81.2\% respectively for the first and the second encoding mechanisms mentioned above. Therefore, the \encabbr{}'s boost in accuracy is due to the unique properties of the proposed components rather than dimension growth.

%% file: secs/exp/action_imp.tex
\subsection{Improvement Analysis}
To get insights into the working mechanisms of \encabbr{} and CRL, we quantitatively compare each action's recognition accuracy before and after applying the two proposed accessories respectively to the BKB model. 

For CRL, its design intention is to dispel the confusion between two actions where one is visually like the temporal inverse of the other by explicitly encoding the chronological order of a skeleton sequence. For example, the hand joints of \textit{opening a box} move in the opposite direction along time to those of \textit{folding a paper}. We observe that the chaos in such extensive pairs of actions has been greatly reduced. We list three action examples in \autoref{subtable:CRL_confuse}. Note that as a result of applying CRL, the most confusing action accordingly changes from the ones with the opposite execution direction along time to those with similar motion trajectories, which implies that the previously missing information on temporal evolution has been sufficiently filled. 

For more general actions, CRL also greatly enhances the accuracy of the other actions apart from the above ones, as \autoref{subtable:CRL_improve} exhibits. This phenomenon indicates that information about temporal evolution is universally beneficial for accurately recognizing general actions. Such information is insufficiently captured without CRL. On the other hand, as \autoref{subtable:TTE_improve} depicts, \encabbr{} also broadly boosts the recognizer's accuracy over many actions. We hypothesize this is because most skeleton action sequences contain a level of noise, and our \encabbr{} effectively mitigates its adverse influence on skeleton-based action recognition.

%% file: secs/exp/noise_alleviation.tex
\begin{figure}[t!]
    \centering
    \includegraphics[width=0.6\linewidth]{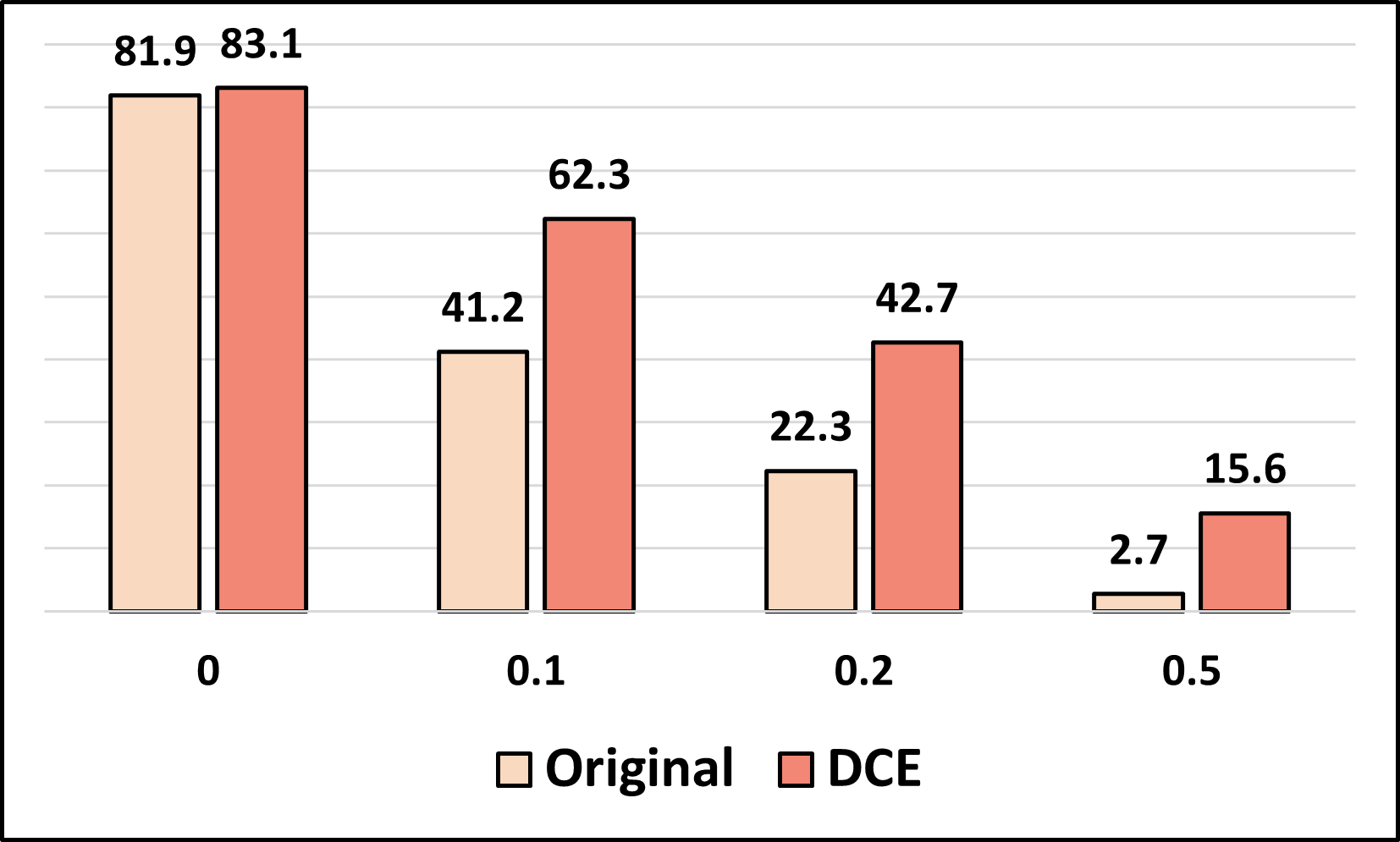}
    \caption{Recognition accuracy on the perturbed validation set with varying noise levels. \encabbr{} stands for encoding the skeleton action sequence with \encabbr{}. }
    \label{fig:noise_alleviation}
\end{figure}

% \vspace{-2mm}
\subsection{Noise Alleviation}
The proposed \encabbr{} decomposes the original skeleton action sequence into different frequency components. We select the lower ones with the intention of alleviating noise in the signal. Here, we empirically verify the effectiveness of noise mitigation. To this end, for each instance in the validation set, we randomly sample noise from a spherical Gaussian distribution: $\mathbf{n} \sim \ngaussian$ with the same shape as the sequence. This is followed by adding it to the sequence as: $\mathbf{x} := \mathbf{x} + \epsilon \cdot \mathbf{n}$, where $\epsilon$ is a hyperparameter to manually control the intensity of noise. Larger $\epsilon$ imposes stronger noise. We use the cross-subject setting and the joint features of NTU120. As \autoref{fig:noise_alleviation} shows, we see that using \encabbr{} consistently resists more against the accuracy drop on the perturbed validation set compared with that of not utilizing \encabbr{}, testing on a range of $\epsilon$ values, directly verifying the effectiveness of the proposed \encabbr{}'s noise alleviation.

%% file: secs/exp/compatibility.tex
\subsection{Compatibility with Existing Models}
The proposed \methodabbr{} is not merely applicable to the baseline model. The two components are widely compatible with existing networks. We desire to know the quantitative accuracy that \methodabbr{} improves for an action recognizer. To this end, we train recently proposed models with and without \methodabbr{}. The model pair is fairly trained and compared by setting the same hyperparameters. The utilized dataset is NTU60 cross-subject using the joint features. We summarize the results in \autoref{fig:model_accuracy_bar_chart}. Out of the seven selected action recognizers, \methodabbr{} consistently increases the action recognizer's accuracy, indicating the \methodabbr{}'s great compatibility. 

%% file: secs/exp/sota.tex
\subsection{Comparison with SOTA Accuracy}
We show in \autoref{tab:compare_with_sota} that extending the simple baseline with \methodabbr{} outperforms the current competitive state-of-the-art action recognizers; even the number of parameters and the inference time of the utilized model (BKB) are respectively far smaller and shorter than the SOTA networks. Note that \methodabbr{} is generally compatible with existing models. Hence, replacing the utilized simple backbone with a more complicated model can achieve even higher accuracy. To illustrate, we equip MSG3D with \methodabbr{} and obtain the accuracy of \textbf{87.5\%} and \textbf{89.1\%}, respectively on the cross-subject and cross-setup of NTU120. As a comparison, the original MSG3D's accuracy results on these two settings without using the proposed components are 86.9\% and 88.4\%, respectively. Similarly, this combination achieves \textbf{91.8\%} and \textbf{96.8\%} on the cross-subject and cross-view of NTU60, achieving new SOTA accuracy.

%% file: secs/conclusion.tex
\vspace{-1mm}
\section{Conclusion}
\vspace{-1mm}
We propose two temporal accessories that are compatible with existing skeleton-based action recognizers for boosting their accuracy. On the one hand, discrete cosine encoding (\encabbr) facilitates a model to comprehensively capture information from the frequency domain. Additionally, we can proactively alleviate signal noise by highlighting the low-frequency components of an action sequence. On the other hand, the proposed chronological loss directly leads a network to be aware of the action sequence's temporal direction. Our experimental results on benchmark datasets show that incorporating the two components consistently improves existing skeleton-based action recognizers' performance.  